\documentclass[]{article}  

\usepackage{amsmath,amsfonts,amssymb}
\usepackage{graphicx}
\usepackage[colorlinks=true, allcolors=blue]{hyperref}
\usepackage{subcaption}
\usepackage{adjustbox}
\usepackage{booktabs}
\usepackage{multirow}
\usepackage{arydshln}      

\usepackage{authblk} 
\usepackage[a4paper,margin=25mm]{geometry}

\usepackage{algorithm}
\usepackage{algpseudocode}

\title{Neural Implicit 3D Cardiac Shape Reconstruction from Sparse CT Angiography Slices Mimicking 2D Transthoracic Echocardiography Views}

\author[1,2]{Gino E. Jansen\thanks{Correspondence: \href{mailto:g.e.jansen@amsterdamumc.nl}{g.e.jansen@amsterdamumc.nl}}}
\author[1,2]{Carolina Brás}
\author[3]{R. Nils Planken}
\author[4]{Mark J. Schuuring}
\author[5]{Berto J. Bouma}
\author[1,3,6]{Ivana Išgum}
\date{}

\affil[1]{Department of Biomedical Engineering \& Physics, Amsterdam UMC, Netherlands}
\affil[2]{qurAI group, Informatics Institute, University of Amsterdam, Netherlands}
\affil[3]{Department of Radiology, Mayo Clinic, Rochester, Minnesota, USA}
\affil[4]{Department of Biomedical Signals and Systems, University of Twente, Netherlands}
\affil[5]{Department of Cardiology, Amsterdam UMC, Netherlands}
\affil[6]{Department of Radiology \& Nuclear Medicine, Amsterdam UMC, Netherlands}

\begin{document} 
\maketitle

\begin{abstract}
Accurate 3D representations of cardiac structures allow quantitative analysis of anatomy and function. 
In this work, we propose a method for reconstructing complete 3D cardiac shapes from segmentations of sparse planes in CT angiography (CTA) for application in 2D transthoracic echocardiography (TTE). 
Our method uses a neural implicit function to reconstruct the 3D shape of the cardiac chambers and left-ventricle myocardium from sparse CTA planes. 
To investigate the feasibility of achieving 3D reconstruction from 2D TTE, we select planes that mimic the standard apical 2D TTE views.
During training, a multi‐layer perceptron learns shape priors from 3D segmentations of the target structures in CTA. 
At test time, the network reconstructs 3D cardiac shapes from segmentations of TTE-mimicking CTA planes by jointly optimizing the latent code and the rigid transforms that map the observed planes into 3D space. 
For each heart, we simulate four realistic apical views, and we compare reconstructed multi‐class volumes with the reference CTA volumes. 
On a held‐out set of CTA segmentations, our approach achieves an average Dice coefficient of 0.86 ± 0.04 across all structures. 
Our method also achieves markedly lower volume errors than the clinical standard, Simpson’s biplane rule: 4.88 ± 4.26 mL vs. 8.14 ± 6.04 mL, respectively, for the left ventricle; and 6.40 ± 7.37 mL vs. 37.76 ± 22.96 mL, respectively, for the left atrium.
This suggests that our approach offers a viable route to more accurate 3D chamber quantification in 2D transthoracic echocardiography.

\end{abstract}

\section{Introduction}

Despite the availability of 3D transthoracic echocardiography (TTE), its limited spatiotemporal resolution means that 2D acquisitions remain the standard in clinical practice. 
Key clinical assessments include the estimation of chamber size and function, which require geometric assumptions to approximate volumes from planes -- such as Simpson’s biplane rule~\cite{kircher1991left, lang2015recommendations} for the apical 2- and 4-chamber views. 
With such approximations, however, misalignment of the probe with the apex of the heart can cause systematic underestimation of chamber volumes~\cite{kim2022beyond}. 
A method that infers the 3D position of the ultrasound planes and reconstructs the 3D cardiac anatomy could improve volumetric accuracy.

Several authors addressed 3D cardiac shape reconstruction from 2D echocardiography, where estimation of the view pose -- the orientation of the acquired plane -- has been crucial\cite{freitas2024automatic, laumer20252d, chen20253d}. The methods used deep-learning pose estimation with statistical shape models \cite{freitas2024automatic}, graph neural networks~\cite{laumer20252d}, and neural implicit functions~\cite{chen20253d}. 
{Freitas et al. \cite{freitas2024automatic} first created a deep learning model to perform view-pose estimation using synthetic 3D cardiac shapes of all four chambers, and subsequently, applied statistical shape modeling to find a matching 3D volume \cite{ravikumar2018group}. 
Laumer et al. \cite{laumer20252d} proposed a single model to jointly find the view pose and reconstruct the left-ventricle shape using a test-time optimization scheme for a graph-neural network. Their pipeline iteratively updated the view poses and cardiac shape based on a loss that measured the similarity between 2D-rendered cross-sections of the 3D shape and 2D segmentations from TTE. 
Chen et al.\cite{chen20253d} proposed a test-time optimization scheme with a neural implicit function, updating view poses and the shape of a single left- and right-ventricle template in an alternating fashion.
While demonstrating successful cardiac shape reconstructions, the test-time optimization schemes required all of the weights of the model to be updated. These approaches therefore risk catastrophic forgetting of the learned prior\cite{kirkpatrick2017overcoming}. Additionally, all previous approaches used synthetic shape datasets as shape priors, which may not accurately represent the variation present in cardiac shapes from real patients.}

In this work, we use high-resolution shapes from real computed tomography angiography (CTA) acquisitions, and reconstruct full 3D cardiac anatomy from sparse CTA slices that mimic four apical TTE views. 
We build on previous work by Amiranashvili et al.~\cite{amiranashvili2024learning}, who introduced a general framework for implicit shape reconstruction from sparse slices by learning shape priors from complete 3D shapes -- which, in our work, come from CTA-derived whole-heart segmentations. 
The approach uses a multi-layer perceptron (MLP) to learn from densely segmented 3D shapes; each input coordinate is concatenated with a learnable per-shape latent code, enabling the model to represent a distribution of plausible shapes. 
At test time, the network reconstructs a 3D shape from sparse cross-sections by optimizing only the latent code, keeping network weights fixed.
We extend the method to, at test time, jointly optimize both the latent code and the rigid transformations that map simulated 2D TTE views into 3D space, estimating the true view poses. 
{To simulate the imprecise knowledge of the true poses in free-hand acquisitions, we sample four slices from a 3D cardiac shape representing apical TTE views, and randomly perturb those poses.}
Reconstruction accuracy is assessed on a held-out set of CTA segmentations, comparing multi-class volumes reconstructed by our method against reference CTA volumes and against those obtained with the current clinical standard, Simpson’s biplane rule~\cite{kircher1991left, lang2015recommendations}.

\section{Data}
\label{sec:data}
We used a dataset of 452 patients (median age: 72 years IQR: 62–81, 59.3\% male) with acute ischemic stroke, who were assessed for cardioembolism using cardiac CTA (prospectively ECG-gated, 100 kVp, 288 mAs, in-plane resolution 0.29-0.52 mm, 0.6 mm slice thickness, 0.4 mm increment)~\cite{rinkel2022diagnostic}. Segmentations of the left atrium (LA), left ventricle (LV), right atrium (RA), right ventricle (RV), and left ventricular myocardium (LV-myo) at end-diastole were obtained using a previously developed deep learning model~\cite{Bruns2022DeepCT}. From the full cohort, 153 subjects were selected {based on qualitative assessment of automatic segmentation quality. No additional post-processing or re-orientation of the segmentation maps was applied -- all the volumes already shared the same anatomical orientation convention and slice dimensions (512 $\times$ 512).}

\begin{figure}[ht!]
    \centering
    \includegraphics[width=\linewidth]{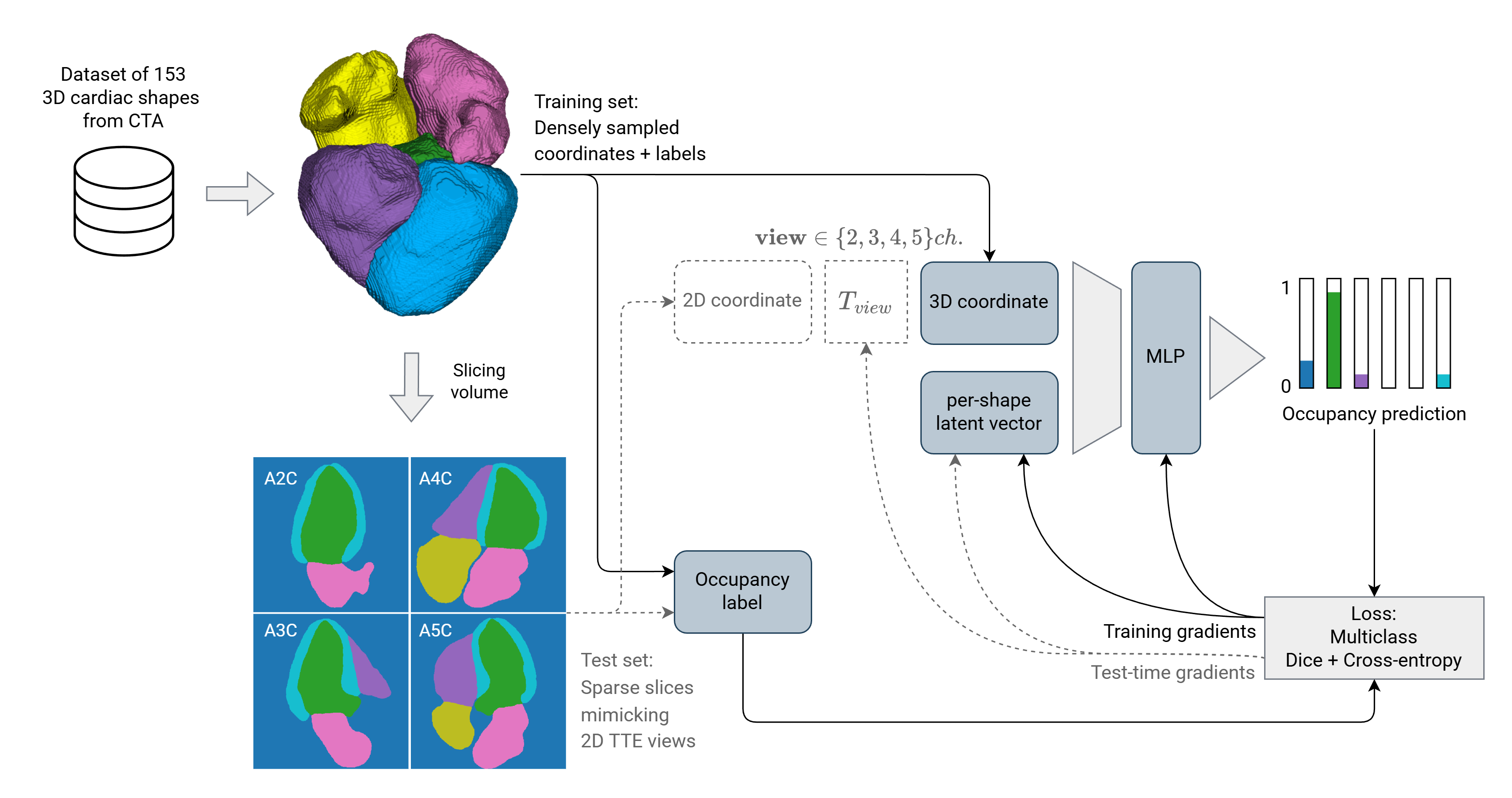}
    \caption{Method overview. Dashed elements represent operations that are exclusive to testing. 
    Training: an MLP learns multi-class occupancies from dense CTA points concatenated with a per-shape latent code. 
    Testing: 2D masks are mapped to 3D via a rigid transform \(T_{\text{view}}\); \(T_{\text{view}}\) and the latent code are optimized while the MLP is frozen.}
    \label{fig:method}
\end{figure}

\section{Method}

Our approach extends the implicit shape reconstruction framework by Amiranashvili et al.~\cite{amiranashvili2024learning} to enable multi-class cardiac structure segmentation from sparse 2D multi-view masks, with simultaneous optimization of both the latent representation and the 2D-to-3D plane mapping at test time. An overview of the proposed methodology is provided in Figure \ref{fig:method}.

\subsection{Learning a shape prior with a neural implicit function}
We adopt the autodecoder-based neural implicit function~\cite{amiranashvili2024learning}. A multi-layer perceptron (MLP) $f_\theta$ is trained to map raw 3D Cartesian coordinates $\mathbf{x} \in \mathbb{R}^3$ {(in mm) } and a latent vector $\mathbf{z} \in \mathbb{R}^{128}$ to per-voxel class occupancies:
$
f_\theta(\mathbf{x}, \mathbf{z}) = \mathbf{p} \in [0,1]^6,
$
where $\mathbf{p}$ is a softmax probability vector over six classes (background + 5 structures). The network architecture consists of 8 hidden layers of width 128, with skip connections as described in~\cite{amiranashvili2024learning}. Each subject in the training set is assigned a unique latent code, jointly optimized with the network parameters.
The network is trained using a combination of categorical cross-entropy and multi-class soft Dice loss:
$
\mathcal{L}_\text{data} = \mathcal{L}_\text{CE} + \mathcal{L}_\text{Dice}
$
with an additional $L_2$-regularization term on the latent codes:
$
\mathcal{L} = \mathcal{L}_\text{data} + \lambda \|\mathbf{z}\|_2^2, 
$
where $\lambda = 1 \times 10^{-4}$. We use Adam optimizer with an initial learning rate of $1 \times 10^{-4}$ for the network parameters and latent codes. Training is performed for 1,800 epochs with a batch size of 8 and 64$^3$ randomly sampled points per volume per iteration. 

\subsection{Slicing 2D TTE-mimicking views from CTA segmentations}
\label{sec:view-slice}
{To simulate 2D transthoracic echocardiography (TTE) views for testing, we automatically sample multi-class segmentation slices mimicking apical 2-, 3-, 4-, and 5-chamber (A2C, A3C, A4C, A5C) planes from the 3D CTA segmentations. We first compute the chambers' centers of mass (CoMs) from the 3D segmentations and calculate the apex location as the LV voxel farthest from the LA CoM. Each plane is then parameterized by an anchor point $\mathbf{a}\in\mathbb{R}^3$ and two in-plane basis vectors $\mathbf{e}_u,\mathbf{e}_v\in\mathbb{R}^3$. We set $\mathbf{a}$ to the apex; define $\mathbf{e}_v$ as the (unit) long-axis direction from the apex to the LV CoM; and obtain the (unit) short-axis direction $\mathbf{e}_u$ by projecting the vector from apex to RA CoM onto the plane orthogonal to $\mathbf{e}_v$ (and normalizing to unit length). Using this canonical basis, we define the A4C plane, which is spanned by $(\mathbf{e}_u,\mathbf{e}_v)$. Then, we generate the A3C and A2C by rotating $\mathbf{e}_u$ about $\mathbf{e}_v$ by $45^{\circ}$ and $90^{\circ}$, respectively, and the A5C by a $5^{\circ}$ superior tilt of the A4C (rotation about $\mathbf{e}_u$). For A4C and A5C, we additionally apply a $10^{\circ}$ in-plane rotation to center the full heart in the field-of-view. To render a 2D image, we sample a grid of pixels $(i,j)$, with the 3D coordinates of each pixel defined as a linear combination of the basis vectors:
\begin{equation}
    \label{eq:plane}
    \mathbf{x}_{ij}=\mathbf{a}+\alpha_{ij}\mathbf{e}_u+\beta_{ij}\mathbf{e}_v,
\end{equation}
where $(\alpha_{ij},\beta_{ij})$ represent the local 2D plane coordinates.
Each pixel is assigned the nearest-neighbor label from the 3D segmentation at $\mathbf{x}_{ij}$. We extract $256\times256$ images with an extent of $1.5\times|\mathbf{a}-\mathrm{CoM}_{\mathrm{LA}}|$ in both directions; the short-axis extent is symmetric around zero, while the long-axis grid ranges from $-0.1$ to $0.9$ of the extent (placing the apex near the top of the image, see Figure~\ref{fig:method}).}

\subsection{Test-time optimization of shape latent and view poses}
At inference, we extend the test-time optimization procedure of Amiranashvili et al.~\cite{amiranashvili2024learning} by jointly optimizing two sets of parameters: (i) the latent code $\mathbf{z}$, and (ii) the parameters of a rigid transformation applied to each canonical 2D plane. 
Optimization of the latter parameter set creates robustness to incorrect initial estimates of the true view poses.
Each view is parameterized by an axis-angle rotation vector $\boldsymbol{\alpha}\in\mathbb{R}^3$ and a translation $\mathbf{t}\in\mathbb{R}^3$, initialized to zero. {Concretely, denoting the canonical plane by anchor point $\mathbf{a}$ and in-plane basis vectors $\mathbf{e}_u,\mathbf{e}_v$, we update the 3D coordinates of each pixel $(i,j)$ (Equation \ref{eq:plane}) as
\begin{equation}
\mathbf{x}_{ij}(\boldsymbol{\alpha},\mathbf{t})=(\mathbf{a}+\mathbf{t})+\alpha_{ij},\mathbf{R}(\boldsymbol{\alpha})\mathbf{e}_u+\beta_{ij},\mathbf{R}(\boldsymbol{\alpha})\mathbf{e}_v,
\label{eq:plane_rigid}
\end{equation}
where rotation matrix $\mathbf{R}(\boldsymbol{\alpha})\in SO(3)$ is obtained from the axis--angle vector using Rodrigues' formula.}
The objective minimizes the sum of cross-entropy and multi-class Dice loss between the network predictions and the observed 2D segmentation masks, backpropagating gradients through both the latent vector and the rigid transformation parameters. {During inference we employ the same $L_2$-regularization on $\mathbf{z}$ as used during training.}

Optimization proceeds in two phases: for the first 100 steps, we only update the latent vector, while the view transformation parameters remain fixed. Subsequently, we optimize all parameters jointly for a total of 1,000 steps using Adam with a learning rate of $1 \times 10^{-2}$. We impose no additional regularization on the transformation parameters $\boldsymbol{\alpha}$ and $\mathbf{t}$. {Once optimized, we create a dense reconstruction of the shape by querying a 3D grid of coordinates from the network that equals the coordinate grid of the reference volume. For every voxel, we determine the occupancy value as the argmax of the multi-class network output.}

\section{Experiments \& Results}

From the set of 153 patients, we trained the proposed method on cardiac shapes from 100 patients, used 13 patients for hyperparameter-tuning, and finally evaluated our method on a test set of 40 patients. {We conducted four experiments: (i) a realistic setting with perturbed view initialization and joint latent-and-pose test-time optimization, (ii) an ablation study that disabled pose optimization, (iii) an idealized setting with true (unperturbed) poses to establish an upper bound, and (iv) a comparison against Simpson’s biplane method using only apical 2-chamber (A2C) and 4-chamber (A4C) views.}

\setlength{\tabcolsep}{2pt} 
\renewcommand{\arraystretch}{1.5}
\begin{table*}[t!]
\centering
\caption{Reconstruction performance in comparison with reference CTA segmentations (mean $\pm$ std) for all experiments and cardiac structures. {For each column, the best result is displayed in bold typeface (i.e., best among experiments above the dashed line).} ASSD = Average symmetric surface distance, MAE = Mean absolute error of chamber volume.}
\renewcommand{\arraystretch}{1.5}

\resizebox{\textwidth}{!}{%
\begin{tabular}{lcccccccc}
\toprule
\multirow{2}{*}{Experiment}
  & \multicolumn{4}{c}{Left ventricle (blood pool)}
  & \multicolumn{4}{c}{Left atrium} \\
\cmidrule(lr){2-5}\cmidrule(lr){6-9}
  & Dice & ASSD (mm) & MAE (mL) & MAE (\%)
  & Dice & ASSD (mm) & MAE (mL) & MAE (\%) \\
\midrule
Latent + pose (perturbed)         
  & \textbf{0.90 $\pm$ 0.03} & \textbf{1.38 $\pm$ 0.55} & \textbf{4.88 $\pm$ 4.26} & \textbf{3.97 $\pm$ 2.57}
  & \textbf{0.87 $\pm$ 0.04} & \textbf{1.88 $\pm$ 0.76} & \textbf{6.40 $\pm$ 7.37} & \textbf{6.66 $\pm$ 7.78} \\

Latent only (perturbed) 
  & 0.84 $\pm$ 0.06 & 2.42 $\pm$ 0.93 & 9.32 $\pm$ 8.80 & 7.41 $\pm$ 5.59
  & 0.73 $\pm$ 0.11 & 4.18 $\pm$ 2.01 & 14.16 $\pm$ 12.38 & 15.02 $\pm$ 13.40 \\

Latent + pose (A2C+A4C)          
  & 0.89 $\pm$ 0.04 & 1.61 $\pm$ 0.62 & 4.90 $\pm$ 4.48 & 4.41 $\pm$ 3.85
  & 0.83 $\pm$ 0.09 & 2.55 $\pm$ 1.46 & 13.41 $\pm$ 18.74 & 13.83 $\pm$ 20.47 \\

Simpson’s (A2C+A4C)            
  &       --        &      --        & 8.14 $\pm$ 6.04 & 7.81 $\pm$ 5.87
  &       --        &      --        & 37.76 $\pm$ 22.96 & 38.92 $\pm$ 21.72 \\
\hdashline
Latent only (ideal poses)     
  & 0.93 $\pm$ 0.02 & 0.89 $\pm$ 0.33 & 2.82 $\pm$ 3.32 & 2.23 $\pm$ 1.70
  & 0.90 $\pm$ 0.03 & 1.37 $\pm$ 0.41 & 4.88 $\pm$ 3.46 & 5.42 $\pm$ 4.78 \\

\bottomrule
\end{tabular}%
}

\vspace{1em}

\resizebox{\textwidth}{!}{%
\begin{tabular}{lcccccccc}
\multirow{2}{*}{Experiment}
  & \multicolumn{4}{c}{Right ventricle}
  & \multicolumn{4}{c}{Right atrium} \\
\cmidrule(lr){2-5}\cmidrule(lr){6-9}
  & Dice & ASSD (mm) & MAE (mL) & MAE (\%)
  & Dice & ASSD (mm) & MAE (mL) & MAE (\%) \\
\midrule
Latent + pose (perturbed)         
  & \textbf{0.87 $\pm$ 0.04} & \textbf{1.95 $\pm$ 0.67} & \textbf{11.31 $\pm$ 13.25} & \textbf{7.18 $\pm$ 7.69}
  & \textbf{0.84 $\pm$ 0.05} & \textbf{2.39 $\pm$ 0.82} & 10.59 $\pm$ 9.02 & 10.46 $\pm$ 8.89 \\

Latent only (perturbed) 
  & 0.81 $\pm$ 0.05 & 2.89 $\pm$ 0.76 & 15.00 $\pm$ 12.40 & 9.81 $\pm$ 7.35
  & 0.75 $\pm$ 0.08 & 3.90 $\pm$ 1.38 & \textbf{9.48 $\pm$ 8.05} & \textbf{9.76 $\pm$ 8.64} \\

Latent + pose (A2C+A4C)          
  & 0.82 $\pm$ 0.07 & 2.84 $\pm$ 1.24 & 27.14 $\pm$ 33.63 & 16.32 $\pm$ 17.34
  & 0.79 $\pm$ 0.07 & 3.20 $\pm$ 1.13 & 13.92 $\pm$ 13.41 & 13.08 $\pm$ 11.67 \\

\hdashline
Latent only (ideal poses)      
  & 0.90 $\pm$ 0.02 & 1.45 $\pm$ 0.35 & 6.58 $\pm$ 5.56 & 4.22 $\pm$ 3.34
  & 0.87 $\pm$ 0.03 & 1.89 $\pm$ 0.56 & 7.18 $\pm$ 6.55 & 6.71 $\pm$ 5.64 \\
  
\bottomrule
\end{tabular}%
}

\vspace{1em}

\resizebox{0.58\textwidth}{!}{%
\begin{tabular}{lcccc}
\multirow{2}{*}{Experiment}
  & \multicolumn{4}{c}{Left ventricle (myocardium)} \\
\cmidrule(lr){2-5}
  & Dice & ASSD (mm) & MAE (mL) & MAE (\%) \\
\midrule
Latent + pose (perturbed)
  & \textbf{0.82 $\pm$ 0.06} & \textbf{1.35 $\pm$ 0.48} & 6.91 $\pm$ 7.68 & 4.95 $\pm$ 4.21 \\
Latent only (perturbed)
  & 0.71 $\pm$ 0.09 & 2.13 $\pm$ 0.64 & 7.02 $\pm$ 6.72 & 5.25 $\pm$ 4.59 \\

Latent + pose (A2C+A4C)
  & 0.79 $\pm$ 0.07 & 1.54 $\pm$ 0.51 & \textbf{5.24 $\pm$ 4.82} & \textbf{3.87 $\pm$ 3.35} \\

\hdashline
Latent only (ideal poses)
  & 0.88 $\pm$ 0.03 & 0.85 $\pm$ 0.26 & 3.04 $\pm$ 2.79 & 2.25 $\pm$ 1.86 \\
  
\bottomrule
\end{tabular}%
}

\label{tab:results_full}
\end{table*}

\begin{figure}[t]
    \centering
    \includegraphics[width=0.7\linewidth, trim={0 -0.5cm 0 0}]{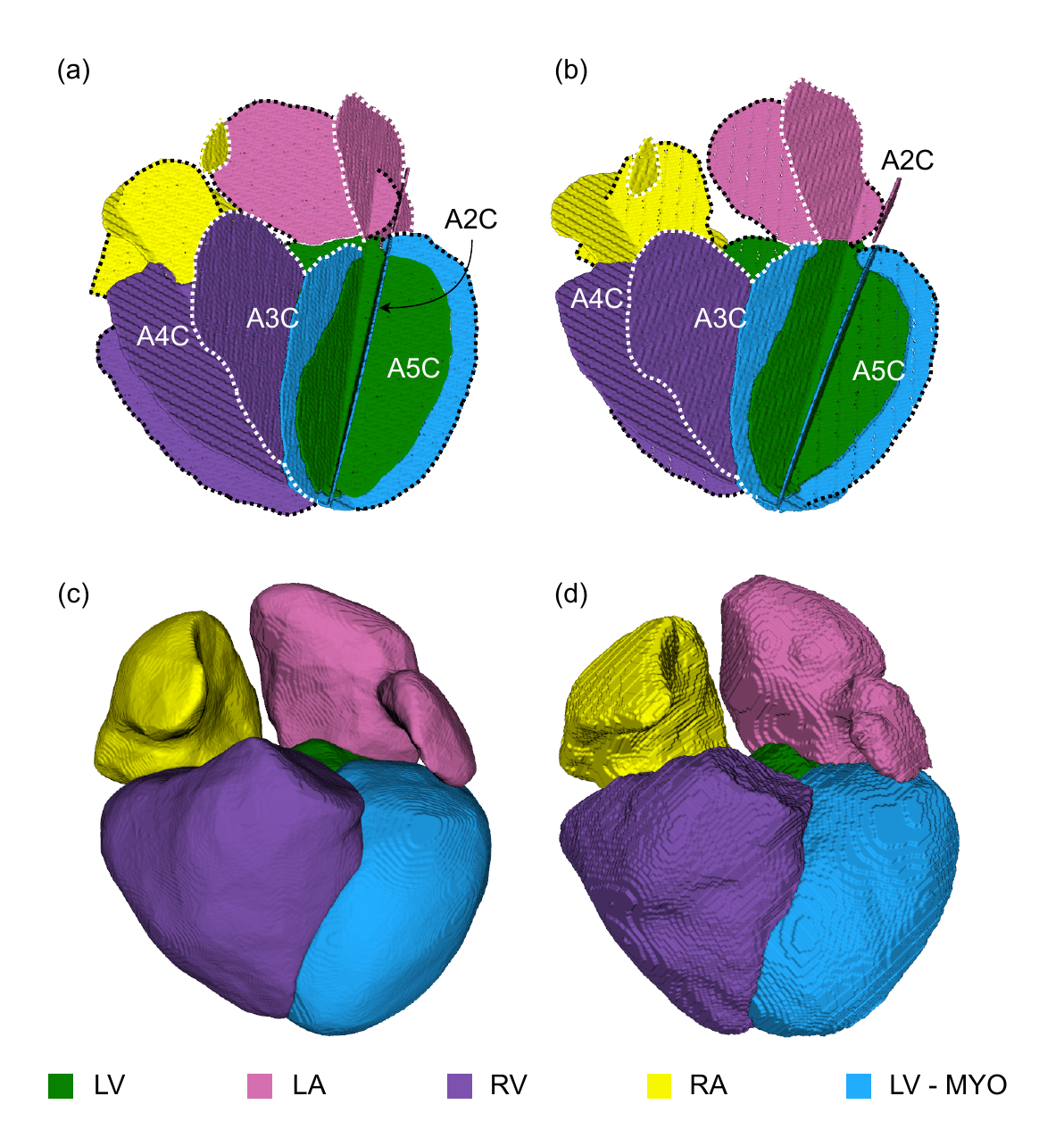}
    \caption{{Example test result. (a) Initial orientation of TTE-mimicking slices showing that the outer borders of the LA in the A3C (white dotted line) and A5C view (black dotted line) are misaligned.
    (b) Slices with optimized orientations, demonstrating improved alignment of the LA borders. (c) Our 3D shape reconstruction from sparse slices shown in (b). (d) 3D rendering of reference CTA segmentation. LV = left ventricle, LA = left atrium, RV = right ventricle, RA = right atrium, LV-MYO = left-ventricle myocardium.}}
    \label{fig:results_A}
\end{figure}

\begin{figure}[t!]
    \centering
    \includegraphics[width=\linewidth, trim={0 -2cm 0 0}]{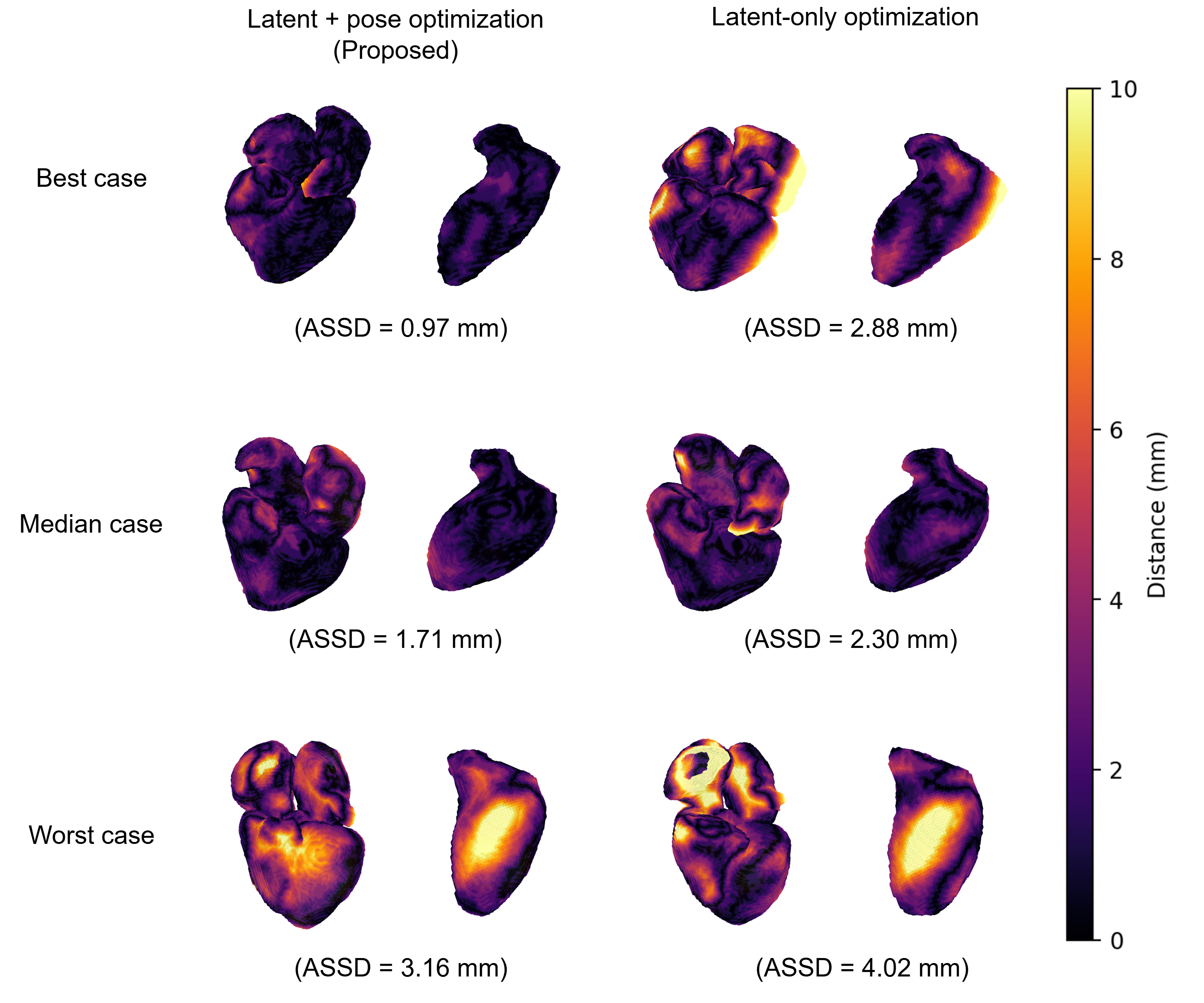}
    \caption{{Heatmap visualizations of surface distance errors of reconstructed whole-heart and left-ventricle shapes. Errors are computed between the reconstructed shape and reference CTA segmentation. In this comparison between two experiments, the cases are determined based on the average symmetric surface distance (ASSD) of the proposed method (left).}}
    \label{fig:results_B}
\end{figure}

\subsection{Shape latent + view pose optimization with perturbed initialization}

The true orientations of real TTE acquisitions are known only approximately by the displayed anatomy that defines a view. 
Therefore, to simulate the imprecise true orientation of a view, we applied Gaussian perturbations ($\sigma = 5.0$ mm) to the landmarks that we used to automatically slice the heart. 
These landmarks included the apex and centers of mass of the cardiac chambers (see Section \ref{sec:view-slice}).
For each view, we re-sampled Gaussian noise and applied it to the landmarks before sampling the corresponding slice.
Subsequently, we replaced the pose parameters of the acquired (noisy) views ($\mathbf{e}_u$,$\mathbf{e}_v$, and $\mathbf{a}$) with the unperturbed (ideal) pose parameters, which represent a realistic initialization.
We utilized the resulting coordinate-label pairs to optimize both the latent code and the plane parameters, resulting in joint reconstruction of a cardiac shape, and alignment of the views with the reconstructed shape.
{The A4C view remained anchored to its true (but perturbed) orientation, and its plane parameters remained fixed. We enforced this to prevent global drift of the reconstructed volume, allowing direct computation of the Average Symmetric Surface Distance (ASSD) and Dice coefficient without the need to realign the reconstruction with the reference.} After optimization, we queried a 3D coordinate grid matching that of the reference volume to obtain the predicted shape reconstruction. From this, we computed Dice, ASSD, and absolute volume errors with respect to the reference. The results are given in Table~\ref{tab:results_full} (\textit{Latent + pose (perturbed)}). Among the reconstructed cardiac structures, the lowest errors are found for the left ventricle. An example shape reconstruction is shown in Figure \ref{fig:results_A}.

\subsection{Shape latent-only optimization with perturbed initialization}
\label{sec:latent-only optimization}

We assessed the contribution of pose optimization to the reconstruction performance in an ablation experiment. 
For this, we repeated our previous experiment, but kept the view pose parameters fixed during optimization, allowing only latent code optimization. Compared to the joint optimization approach, this resulted in higher reconstruction errors, particularly for the LA, which demonstrated an increase in mean volume error from 6.40 mL to 14.16 mL. The result of this experiment is given in Table~\ref{tab:results_full} (\textit{Latent only (perturbed)}). Heatmap visualizations of surface distance errors are given in Figure~\ref{fig:results_B}, showing that optimizing only the shape latent results in larger errors.

\subsection{Upper bound: shape latent-only optimization with ideal initialization}
\label{sec:ideal pose}
To identify an upper bound on the achievable reconstruction accuracy with the current model, we simulated an idealized condition: We acquired the views using the unperturbed landmarks, and initialized their poses to their true (and ideal) values. Only latent codes were optimized in this scenario. This resulted in the highest Dice scores and lowest errors, as shown in Table~\ref{tab:results_full} (\textit{Latent only (ideal poses)}).

\subsection{Comparison to Simpson's biplane method}

We compared our proposed method with the current standard for LA and LV volume estimation -- Simpson’s biplane method --  which utilizes only the A2C and A4C views. For a fair comparison, we ran our proposed method with only those two views during test-time optimization. {We computed biplane Simpson volumes by sampling 20 equally spaced chords along an automatically defined long axis (mitral-plane midpoint to apex) in both A2C and A4C masks, yielding diameters $\{d_{\mathrm{A2C},k}\}$ and $\{d_{\mathrm{A4C},k}\}$. The volume was estimated by the method of disks,
$V=\sum_{k=1}^{N}\frac{\pi}{4}\,d_{\mathrm{A2C},k}\,d_{\mathrm{A4C},k}\,\Delta l$,
with $\Delta l$ the long-axis step size (converted to mL).}

The results of this experiment are given in Table~\ref{tab:results_full}, (\textit{Latent + pose (A2C + A4C)}, and \textit{Simpson's (A2C + A4C)}, respectively). These results show that our proposed model substantially improved volumetric estimates over Simpson’s rule, reducing the mean absolute volume errors for both the LV and LA.

\section{Discussion \& Conclusion}

This study demonstrates that a neural implicit function conditioned on CT‑derived multi‑class shape priors can reliably reconstruct full 3D cardiac anatomy from four slices that mimic the apical 2D TTE views. By jointly optimizing a per‑shape latent vector and the rigid pose parameters of each TTE-mimicking plane at inference, the network achieved a left‑ventricular volume error to below 5\% on average, improving on the current standard for LV volume quantification -- Simpson's biplane method. 

Performance remained robust when reconstruction was limited to the A2C and A4C views, outperforming the Simpson’s method while utilizing the same two views. The latent-only optimization experiment (Section~\ref{sec:latent-only optimization}), further showed that inaccurate assumptions about the plane orientation are a major source of error and that the proposed pose‑optimization strategy mitigates this. However, some residual pose errors persisted. The ideal‑pose experiment (Section~\ref{sec:ideal pose}) indicates algorithmic improvements could be considered, for example through stronger regularization or alternating optimization schedules.

Furthermore, all experiments relied on planes generated from CTA segmentations; the method’s performance on real-world echocardiography scans remains to be established. 
The current method assumes optimal upstream view classification, temporal alignment, and segmentation of the planes, yet each of these automated steps may introduce additional errors and uncertainties that were not addressed here. {Especially automated segmentation in 2D TTE remains challenging, and most widely used benchmarks and methods focus on left-heart structures such as the LV (and, to a lesser extent, the LA)~\cite{leclerc2019deep, ouyang2020video}. The right heart is often poorly resolved in the standard apical views, and benefits from RV-focused acquisitions~\cite{mukherjee2025guidelines}, which are less common in routine TTE examinations.}

{Finally, we fixed the A4C view to its true pose in all experiments, which prevented global drift and allowed overlap metrics to be computed directly against the reference volume. While this choice improved the interpretability of the experiments, it slightly reduced experimental realism by providing one perfectly aligned view.}

{In conclusion, we presented a novel method for 3D cardiac shape reconstruction from sparse 2D segmentations of the heart mimicking 2D transthoracic echocardiography views. For this, we used a neural implicit function that learned a shape prior of the heart from cardiac CT segmentations. Additionally, we integrated view-pose optimization into the test-time optimization framework, which further improved reconstruction quality.}
The proposed implicit reconstruction method provides a promising route towards improved 3D cardiac chamber analysis from routinely acquired 2D TTE, potentially enabling more accurate and reliable assessment of cardiac chamber anatomy and function. 


\section{Acknowledgments}
This work was supported by Pie Medical Imaging BV and Esaote SpA.

\bibliography{report} 
\bibliographystyle{spiebib} 

\end{document}